\documentclass[
]{ceurart}

\usepackage{listings}
\usepackage{graphicx}
\usepackage{array}
\usepackage{hyperref}
\begin{document}

\copyrightyear{2024}
\copyrightclause{Copyright for this paper by its authors.
  Use permitted under Creative Commons License Attribution 4.0
  International (CC BY 4.0).}

\conference{CLEF 2024: Conference and Labs of the Evaluation Forum, September 09–12, 2024, Grenoble, France}

\title{Nullpointer at CheckThat! 2024: Identifying Subjectivity from Multilingual Text Sequence}


\author[1]{Md. Rafiul Biswas}[%
orcid=0000-0002-5145-1990,
email=rafiulbiswas@gmail.com
]
\fnmark[1]
\address[1]{Hamad Bin Khalifa University, Doha, Qatar}
\author[2]{Abrar Tasneem Abir}[%
orcid=0000-0002-2375-4134,
email=abir@cmu.edu,
]
\fnmark[1]
\address[2]{Carnegie Mellon University in Qatar, Education City, Doha, Qatar }
\fnmark[1]
\address[3]{Northwestern University in Qatar, Education City, Doha,
Qatar}

\author[3]{Wajdi Zaghouani}[%
orcid=0000-0003-1521-5568,
email=wajdi.zaghouani@northwestern.edu,
]
\fnmark[1]
\cormark[1]
\cortext[1]{Corresponding author.}

\begin{abstract}
This study addresses a binary classification task to determine whether a text sequence, either a sentence or paragraph, is subjective or objective. The task spans five languages—Arabic, Bulgarian, English, German, and Italian—along with a multilingual category. Our approach involved several key techniques. Initially, we preprocessed the data through parts of speech (POS) tagging, identification of question marks, and application of attention masks. We fine-tuned the sentiment-based Transformer model 'MarieAngeA13/Sentiment-Analysis-BERT' on our dataset. Given the imbalance with more objective data, we implemented a custom classifier that assigned greater weight to objective data. Additionally, we translated non-English data into English to maintain consistency across the dataset. Our model achieved notable results, scoring top marks for the multilingual dataset (Macro F1-0.7121) and German (Macro F1-0.7908). It ranked second for Arabic (Macro F1-0.4908) and Bulgarian (Macro F1-0.7169), third for Italian (Macro F1-0.7430), and ninth for English (Macro F1-0.6893).
\end{abstract}


\maketitle
\section{Introduction}
The concepts of objectivity and subjectivity are crucial in shaping methodologies, interpretations, and the perceived validity of findings in many natural language processing (NLP) applications, such as sentiment analysis and information extraction \cite{wilson2005opinionfinder,gelman2017beyond}. Objectivity analysis relies on data that can be measured, observed, and verified by others and is achieved through careful experimental designs, standard procedures, and statistical analysis. In an ideal sense, objective analysis is supposed to be free from individual biases, emotions, and personal judgments, thereby ensuring that the results are universally valid and replicable \cite{hackett1984decline}. 

Subjectivity, on the other hand, refers to perspectives, interpretations, or analyses that are influenced by personal experiences, feelings, beliefs, or biases \cite{kocon2021learning}. Subjective analysis is inherently shaped by the individual's background, cultural context, and personal viewpoints. While often perceived as less reliable or credible in scientific contexts, subjectivity is an unavoidable aspect of human cognition and can provide valuable insights, particularly in fields such as humanities, social sciences, and qualitative research where personal interpretation and contextual understanding are essential \cite{muller2006subjectivity}.

Identifying whether a text sequence expresses personal opinions, emotions, or factual information is essential for enhancing the accuracy and relevance of automated systems in diverse fields such as social media monitoring, customer feedback analysis, and news content categorization. In data analysis, the tension that arises from the interaction of objectivity and subjectivity frequently affects decision-making procedures and the dissemination of findings. The challenge lies in creating systems that can accurately classify text sequences—whether sentences or paragraphs—as either subjective, reflecting personal opinions or sentiments, or objective, presenting factual information devoid of personal bias \cite{othman2015using}. In an effort to improve the acceptability and credibility of work, researchers may strive for objectivity, occasionally avoiding or hiding choices that might be viewed as subjective. Subjective opinions can, for example, slightly skew the analysis's ostensibly objective results in the data selection, analytical method selection, and result interpretation processes. Thus, there is a high chance that the dataset contains a relatively higher number of objective values compared to subjective values. 

Task 2 in CheckThat Lab at CLEF 2024 \cite{galassi2023overview} classifies text as either subjective or objective. This binary classification task requires systems to accurately identify the nature of a text sequence. The task is offered in multiple languages: Arabic, Bulgarian, English, German, and Italian, providing a comprehensive multilingual evaluation of the systems' capabilities. The challenge of multilingual and cross-linguistic text classification is compounded by the inherent linguistic and cultural differences that influence the expression of subjectivity and objectivity. 

This study presents an approach to a binary classification task aimed at discerning subjective from objective text across multiple languages. By leveraging advanced NLP techniques and Transformer models, we aim to enhance the accuracy and robustness of subjective-objective text classification. The implications of this research extend to improving automated news analysis, enhancing content recommendation systems, and promoting a comprehension understanding in various languages.

\section{Related Works}
The task of classifying text as subjective or objective has been studied extensively in natural language processing. Early work by \citet{wiebe1999development} laid the foundations for subjectivity analysis, proposing a scheme for annotating subjective elements in text. They developed a system called OpinionFinder \citep{wilson2005opinionfinder} which performed subjectivity analysis using various lexical and syntactic features.
More recently, deep learning approaches have been applied to this task with great success. \citet{nakov2016semeval} provide a thorough overview of modern approaches to sentiment analysis, including detecting subjectivity. They highlight the effectiveness of leveraging pre-trained language models like BERT \citep{devlin2019bert} and fine-tuning them for the target task.
Several studies have specifically examined subjectivity classification in a multilingual setting. \citet{balahur2009opinion} constructed a multilingual dataset for subjectivity classification in English, Spanish, French and German. They experimented with various machine translation approaches to make the problem cross-lingual. Similarly, \citet{mihalcea2007learning} generated subjectivity datasets for English and Romanian, using English tools and manually translating the subjective sentences into Romanian.
The CLEF \cite{10.1007/978-3-031-56069-9_62}(Conference and Labs of the Evaluation Forum) has run workshops on automatic identification and verification of claims in political debates, speeches, and news articles since 2018 \citep{nakov2022clef}. The CheckThat! shared task at CLEF focuses on detecting checkworthy claims across various languages including Arabic \citep{alam2021fighting}, which is one of the languages in the current study.
In terms of methodology, fine-tuning pre-trained Transformer models has proven very effective for subjectivity and sentiment tasks. \citet{xu2019bert} fine-tuned BERT for sentiment classification and demonstrated its strong performance on multiple benchmarks. Exploring multi-task learning, \citet{yu2019adapting} showed that jointly learning sentiment and subjectivity through a shared BERT encoder led to improvements on both tasks.

\section{System Overview}
\begin{figure}[t]
    \centering
    \includegraphics[width=0.75\columnwidth]{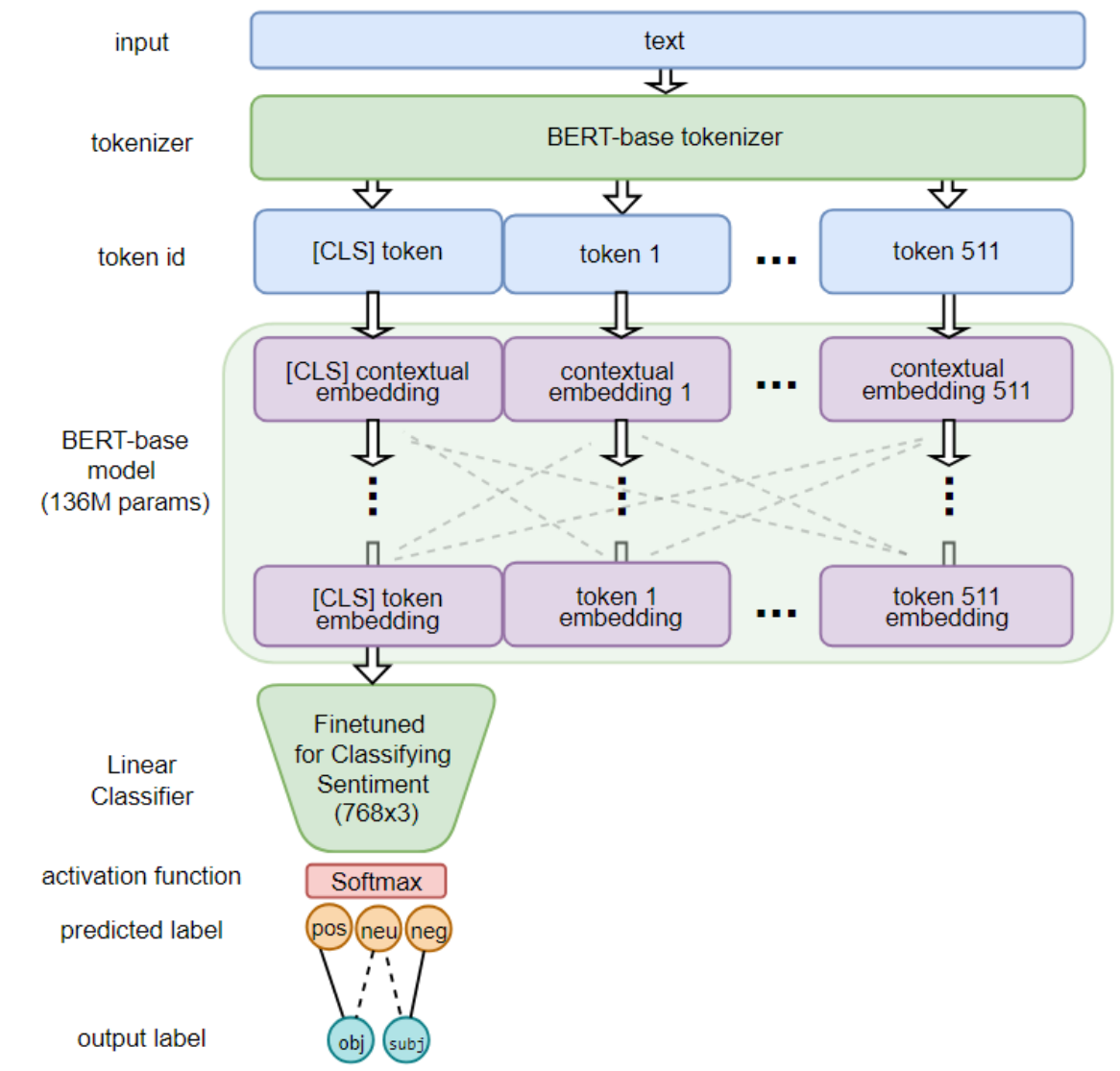}
    \caption{Diagram for classification of subjectivity in text sequence}
    \label{fig:model-development}
\end{figure}

This works system for subjectivity classification comprises several key components, including data preprocessing, model selection, and training strategies. This section provides an overview of each component and the techniques employed (see Figure \ref{fig:model-development}).
\subsection{Data Preprocessing}
The first step in the pipeline is data preprocessing, which involves cleaning and transforming the raw text data into a suitable format for model. The preprocessing steps include:

\begin{itemize}
\item \textbf{Demojization}: We convert emoji characters into their text descriptions using a demojizer to ensure consistent input to the model.
\item \textbf{Removing users and links}: We remove user mentions and URLs from the text, as they are not relevant for subjectivity classification.
\item \textbf{Handling poorly formatted TSV files}: Some of the provided TSV files were poorly formatted, so we use a custom dataset class to handle the processing instead of relying on the pandas library.
\end{itemize}

We also experiment with additional preprocessing techniques such as part-of-speech (POS) tagging and attention masking, but find that they do not significantly improve the performance of the model.
\subsection{Model Selection}
For the subjectivity classification task, we choose to fine-tune pre-trained Transformer models that have been previously trained on sentiment analysis tasks. Specifically, we use the 'MarieAngeA13/Sentiment-Analysis-BERT' model, which is a BERT-based model fine-tuned for sentiment analysis. We find that this approach of using a model already fine-tuned for a related task (i.e., multi-task learning) yields better results compared to fine-tuning a pre-trained model from scratch. The code and data can be found in the GitHub repository \url{https://github.com/Abrar-Abir/CLEF2024task02}.
\subsection{Training Strategies}
The training was conducted on a remote Dell server running the latest Ubuntu 22 OS with 512 GB RAM and 24-core CPU. The server was equipped with NVIDIA A100 GPU with 80 GB GPU memory. We employ several training strategies to improve the performance of the model listed below. 
\begin{itemize}
\item \textbf{Label mapping}: The pre-trained sentiment analysis model is designed for three-class prediction (positive, neutral, negative), while our subjectivity classification task requires only two classes (subjective and objective). We experimented with different label mappings and found that mapping subjective to negative sentiment and objective to positive sentiment yielded the best results.
\item \textbf{Confidence weighting}: For the English dataset, we incorporate the confidence level information provided in the dataset (in the 'solved\_conflict' column where 1[true] means conflict was resolved i.e., higher annotation confidence and vice versa). We assign 20\% higher weight for the training losses- coming from the annotations with higher confidence (i.e, 1.2 weight)- before passing the losses to the loss function so that backpropagation prioritizes minimizing loss for higher confidence annotation compared to their counterparts.
\item \textbf{Hyperparameter tuning}: We experiment with different hyperparameter settings and find that a batch size of 16, learning rate of 2e-5, and training for 20 epochs yields the best performance.
\end{itemize}
\subsection{Language Adaptation}
To handle the multilingual nature of the task, we employ machine translation to convert non-English data into English. We use the Google Translator API through the deep translator library for this purpose. While we also experiment with fine-tuning language-specific pre-trained models for non-English languages, we find that translating the training and test datasets to English and using the English model yields better performance.
These preprocessing, model selection, and training strategies form the core of the subjectivity classification system. In the following sections, we detail our experimental setup and present the results of the approach.

\section{Results}
This section presents the results of subjectivity classification system across various languages and datasets. We first describe the dataset characteristics and then provide a detailed analysis of the model's performance using different evaluation metrics. Finally, we compare our results with those of other participating teams in the CheckThat! Lab at CLEF 2024.

\subsection{Dataset Description}

The dataset for the Subjectivity Subtask consists of sentences from news articles in five languages: Arabic, Bulgarian, English, German, and Italian. Additionally, there is a multilingual dataset that combines all five languages. Table~\ref{tab:dataset} shows the distribution of objective and subjective sentences in the training and test sets for each language. Across all languages, the percentage of objective sentences is higher than that of subjective sentences, with the imbalance being more pronounced in the training sets. This imbalance poses a challenge for subjectivity classification systems, as they need to learn from skewed data distributions.
For the Arabic language, the training set comprises 1185 sentences, with 905 being objective (76.37\verb|%|), and the test set includes 748 sentences, of which  425 are classified as objective (56.81\verb|%|). In Bulgarian, the training set contains 729 sentences, where 406 are objective (55.69\verb|%|), and the test set consists of 250 sentences, with 143 objective sentences (57.2\verb|%|). The English dataset includes 830 sentences for training, with 532 labeled as objective (64.09\verb|%|), and the test set has 484 sentences, with 362 objectives (74.79\verb|%|). For the German language, the training set comprises 800 sentences, of which 492 are objective (61.5\verb|%|), and the test set contains 337 sentences, with 226 being objective (67.07\verb|%|). For the Italian language, the training set includes 1613 sentences, with 1231 objectives (76.31\verb|%|), and the test set comprises 513 sentences, with 377 objectives (73.4\verb|%|). The multilingual dataset combines all five languages and comprises 5159 sentences in the training set, of which 3568 are objective (69.16\verb|%|). The test set contains 500 sentences, evenly split with 250 objective sentences (50\verb|%|) and 250 subjective sentences (50\verb|%|). This comprehensive dataset provides a robust foundation for developing and evaluating systems that distinguish between subjective and objective statements in news articles across multiple languages.

\begin{table}[ht]
\centering
\caption{Training and Test Data Distribution}
\begin{tabular}{llcc}
\toprule
Language    & Dataset       & OBJ (N) (\%) & SUBJ (N) (\%) \\
\midrule
Arabic      & Train (1185)  & 905 (76.37)  & 280 (23.63)   \\
            & Test (748)    & 425 (56.81)  & 323 (43.18)   \\
Bulgarian   & Train (729)   & 406 (55.69)  & 323 (44.23)   \\
            & Test (250)    & 143 (57.2)   & 107 (42.8)    \\
English     & Train (830)   & 532 (64.09)  & 298 (35.9)    \\
            & Test (484)    & 362 (74.79)  & 122 (25.2)    \\
German      & Train (800)   & 492 (61.5)   & 308 (38.5)    \\
            & Test (337)    & 226 (67.07)  & 111 (32.93)   \\
Italian     & Train (1613)  & 1231 (76.31) & 382 (23.68)   \\
            & Test (513)    & 377 (73.4)   & 136 (26.5)    \\
Multilingual& Train (5159)  & 3568 (69.16) & 1591 (30.83)  \\
            & Test (500)    & 250 (50)     & 250 (50)      \\
\bottomrule
\end{tabular}
\label{tab:dataset}
\end{table}

\subsection{Performance Metrics}

We evaluate our subjectivity classification model using various performance metrics, including macro-averaged F1-score, precision, recall, and accuracy. Table \ref{tab:performance_metrics} presents the results for each language and the multilingual dataset. Our model achieves the best performance on the German dataset, with an F1 Macro score of 0.79 and an accuracy of 0.81, indicating high prediction correctness. The multilingual dataset obtains good performance, with an F1 Macro score of 0.71, an F1 SUBJ of 0.69, and an accuracy of 0.71. The model also shows good performance for the Italian language with an F1 Macro score of 0.74 and strong subjective class metrics, with an F1 SUBJ of 0.64. On the other hand, the model struggles the most with the Arabic dataset, obtaining an F1 Macro score of 0.49 and an accuracy of 0.52. The performance is relatively lower than in other languages, which shows the difficulty in identifying subjective data. The model performs well in Bulgarian, achieving an F1 Macro score of 0.72 and high subjective class performance with an F1 SUBJ of 0.69. For English, the performance is moderate to good, with an F1 Macro score of 0.68. The model handles subjective data in English relatively better, with an F1 SUBJ of 0.54, precision (P SUBJ) of 0.52, and recall (R SUBJ) of 0.64. The overall accuracy for English is 0.64.

In summary, the model shows the highest performance in German, followed by Italian and Bulgarian, with Arabic being the most challenging language for the model. The performance in English is moderate, and the overall multilingual performance is strong, suggesting the model's effectiveness across multiple languages but with some variability in specific language performance.

\begin{table}[ht]
    \caption{Performance metrics across different languages}
    \centering
    \begin{tabular}{|c|c|c|c|c|c|c|c|}
        \hline
        Language & F1 Macro & P Macro & R Macro & F1 SUBJ & P SUBJ & R SUBJ & Accuracy \\
        \hline
        Arabic & 0.49 & 0.49 & 0.50 & 0.37 & 0.43 & 0.33 & 0.52 \\
        Bulgarian & 0.72 & 0.72 & 0.72 & 0.69 & 0.66 & 0.72 & 0.72 \\
        English & 0.68 & 0.43 & 0.50 & 0.54 & 0.52 & 0.64 & 0.64 \\
        German & 0.79 & 0.78 & 0.81 & 0.73 & 0.67 & 0.80 & 0.81 \\
        Italian & 0.74 & 0.73 & 0.77 & 0.64 & 0.57 & 0.73 & 0.78 \\
        Multilingual & 0.71 & 0.72 & 0.71 & 0.69 & 0.76 & 0.63 & 0.71 \\
        \hline
    \end{tabular}
    \label{tab:performance_metrics}
\end{table}
\begin{itemize}
    \item \textbf{F1 Macro}: The macro-averaged F1 score, which is the harmonic mean of precision and recall across all classes.
    \item \textbf{P Macro}: The macro-averaged precision.
    \item \textbf{R Macro}: The macro-averaged recall.
    \item \textbf{F1 SUBJ}: The F1 score for subjective classification.
    \item \textbf{P SUBJ}: The precision for subjective classification.
    \item \textbf{R SUBJ}: The recall for subjective classification.
    \item \textbf{Accuracy}: The overall accuracy of the model.
\end{itemize}

\subsection{Comparison with Other Teams}

We compare the performance of our subjectivity classification model with that of other participating teams in the CheckThat! Lab at CLEF 2024. Table \ref{tab:official_results} shows the official results for each language and the multilingual dataset. Our team achieves the highest rank in the German and multilingual categories, with Macro F1 scores of 0.7908 and 0.7121, respectively. We also secure the second position in Arabic and Bulgarian. For Arabic, our model achieved second place with a Macro F1 score of 0.4908 and a SUBJ F1 score of 0.37. In Bulgarian, our model also secured second place with a Macro F1 score of 0.7169 and a SUBJ F1 score of 0.69. In Italian, our model ranks third with a Macro F1 score of 0.7430 and a SUBJ F1 score of 0.64. In the English category, our model ranks ninth with a Macro F1 score of 0.6893 and a SUBJ F1 score of 0.54.

These results showcase the competitiveness of our approach in the shared task, especially in the German and multilingual categories. They also indicate areas for improvement, particularly in English, where our model's performance is lower than other teams. Overall, our team's participation in the ArAIEval shared task demonstrated strong performance across multiple languages, securing top ranks in several categories and showcasing our model's capabilities in multilingual data and subjective data evaluation.

\begin{table}[h]
\caption{Official results for six test languages in Subtask2 CheckThat! Lab at CLEF 2024}
\centering
\begin{tabular}{|c|c|c|c|c|}
\hline
\rowcolor[HTML]{C0C0C0} 
\textbf{Language} & \textbf{Team} & \textbf{Rank} & \textbf{Macro F1} & \textbf{SUBJ F1} \\
\hline
\cellcolor[HTML]{FFFFFF} \textbf{Arabic} & IAI Group & 1 & 0.4947 & 0.46 \\
\cellcolor[HTML]{FFFFFF}  & \textbf{Nullpointer} & 2 & 0.4908 & 0.37 \\
\cellcolor[HTML]{FFFFFF}  & Baseline & 3 & 0.4852 & 0.40 \\
\cellcolor[HTML]{FFFFFF}  & JUNLP (last) & 7 & 0.3623 & 0.00 \\
\hline
\cellcolor[HTML]{FFFFFF} \textbf{Bulgarian} & Baseline & 1 & 0.7531 & 0.73 \\
\cellcolor[HTML]{FFFFFF}  & \textbf{Nullpointer} & 2 & 0.7169 & 0.69 \\
\cellcolor[HTML]{FFFFFF}  & Hybrinfox & 3 & 0.7147 & 0.65 \\
\cellcolor[HTML]{FFFFFF}  & JUNLP (last) & 5 & 0.3639 & 0.00 \\
\hline
\cellcolor[HTML]{FFFFFF} \textbf{English} & Hybrinfox & 1 & 0.7442 & 0.60 \\
\cellcolor[HTML]{FFFFFF}  & \textbf{Nullpointer} & 9 & 0.6893 & 0.54 \\
\cellcolor[HTML]{FFFFFF}  & Baseline & 11 & 0.6346 & 0.45 \\
\cellcolor[HTML]{FFFFFF}  & IAI Group (last) & 15 & 0.4491 & 0.39 \\
\hline
\cellcolor[HTML]{FFFFFF} \textbf{German} & \textbf{Nullpointer} & 1 & 0.7908 & 0.73 \\
\cellcolor[HTML]{FFFFFF}  & IAI Group & 2 & 0.7302 & 0.66 \\
\cellcolor[HTML]{FFFFFF}  & Baseline & 3 & 0.6994 & 0.63 \\
\cellcolor[HTML]{FFFFFF}  & Hybrinfox (last) & 4 & 0.6968 & 0.57 \\
\hline
\cellcolor[HTML]{FFFFFF} \textbf{Italian} & JK\_PCIC\_UNAM & 1 & 0.7917 & 0.69 \\
\cellcolor[HTML]{FFFFFF}  & \textbf{Nullpointer} & 3 & 0.7430 & 0.64 \\
\cellcolor[HTML]{FFFFFF}  & Baseline & 4 & 0.6503 & 0.52 \\
\cellcolor[HTML]{FFFFFF}  & IAI Group (last) & 5 & 0.5862 & 0.49 \\
\hline
\cellcolor[HTML]{FFFFFF} \textbf{Multilingual} & \textbf{Nullpointer} & 1 & 0.7121 & 0.69 \\
\cellcolor[HTML]{FFFFFF}  & Hybrinfox & 2 & 0.6849 & 0.63 \\
\cellcolor[HTML]{FFFFFF}  & Baseline & 3 & 0.6697 & 0.66 \\
\cellcolor[HTML]{FFFFFF}  & IAI Group (last) & 4 & 0.6292 & 0.67 \\
\hline
\end{tabular}
\label{tab:official_results}
\end{table}

\section{Discussion}
Our system leveraged state-of-the-art pre-trained language models, specifically BERT, which we fine-tuned for subjectivity classification task. Through extensive experiments, we demonstrated the effectiveness of our approach, achieving competitive performance in various languages. Our system ranked first in the German and multilingual categories, second in Arabic and Bulgarian, and third in Italian. These results highlight the robustness of our model and its ability to generalize across different languages. We also investigated the impact of various preprocessing techniques, such as part-of-speech tagging and attention masking, on the performance of our system.

Furthermore, our analysis of the dataset characteristics revealed the challenges posed by the imbalance between objective and subjective sentences across all languages. This imbalance underscores the need for developing strategies to handle skewed data distributions effectively.

Our work contributes to the growing body of research on subjectivity classification and multilingual natural language processing. The insights gained from our experiments can inform future research directions and help develop more robust and accurate systems for subjectivity analysis across diverse languages.

However, our study also has some limitations. The performance of our system in English was relatively lower compared to other languages, indicating room for improvement. Future work could explore more advanced techniques, such as domain adaptation and transfer learning, to enhance the model's performance in English and other languages. Moreover, the scope of our study was limited to the dataset provided by the CheckThat! Lab. Further research could investigate the generalizability of our approach to other datasets and domains, such as social media and customer reviews.

\section{Conclusion}
In conclusion, our subjectivity classification system, Nullpointer, demonstrates the potential of leveraging pre-trained language models and multilingual approaches for identifying subjective and objective statements in news articles. As the volume of online content continues to grow, the ability to automatically distinguish between subjective and objective information becomes increasingly crucial. Our work contributes to this important research area and paves the way for more advanced and reliable subjectivity analysis systems in the future.

\section{Acknowledgments}
We acknowledge Qatar
National Research Fund grant NPRP14C0916-210015 from the Qatar Research Development and Innovation Council (QRDI) for funding this research.

\begin{thebibliography}{17}
\expandafter\ifx\csname natexlab\endcsname\relax\def\natexlab#1{#1}\fi
\providecommand{\url}[1]{\texttt{#1}}
\providecommand{\href}[2]{#2}
\providecommand{\path}[1]{#1}
\providecommand{\DOIprefix}{doi:}
\providecommand{\ArXivprefix}{arXiv:}
\providecommand{\URLprefix}{URL: }
\providecommand{\Pubmedprefix}{pmid:}
\providecommand{\doi}[1]{\href{http://dx.doi.org/#1}{\path{#1}}}
\providecommand{\Pubmed}[1]{\href{pmid:#1}{\path{#1}}}
\providecommand{\bibinfo}[2]{#2}
\ifx\xfnm\relax \def\xfnm[#1]{\unskip,\space#1}\fi
\bibitem[{Wilson et~al.(2005)Wilson, Hoffmann, Somasundaran, Kessler, Wiebe, Choi, Cardie, Riloff, and Patwardhan}]{wilson2005opinionfinder}
\bibinfo{author}{T.~Wilson}, \bibinfo{author}{P.~Hoffmann}, \bibinfo{author}{S.~Somasundaran}, \bibinfo{author}{J.~Kessler}, \bibinfo{author}{J.~Wiebe}, \bibinfo{author}{Y.~Choi}, \bibinfo{author}{C.~Cardie}, \bibinfo{author}{E.~Riloff}, \bibinfo{author}{S.~Patwardhan},
\newblock \bibinfo{title}{Opinionfinder: A system for subjectivity analysis},
\newblock in: \bibinfo{booktitle}{Proceedings of HLT/EMNLP 2005 Interactive Demonstrations}, \bibinfo{year}{2005}, pp. \bibinfo{pages}{34--35}.
\bibitem[{Gelman and Hennig(2017)}]{gelman2017beyond}
\bibinfo{author}{A.~Gelman}, \bibinfo{author}{C.~Hennig},
\newblock \bibinfo{title}{Beyond subjective and objective in statistics},
\newblock \bibinfo{journal}{Journal of the Royal Statistical Society Series A: Statistics in Society} \bibinfo{volume}{180} (\bibinfo{year}{2017}) \bibinfo{pages}{967--1033}.
\bibitem[{Hackett(1984)}]{hackett1984decline}
\bibinfo{author}{R.~A. Hackett},
\newblock \bibinfo{title}{Decline of a paradigm? bias and objectivity in news media studies},
\newblock \bibinfo{journal}{Critical Studies in Media Communication} \bibinfo{volume}{1} (\bibinfo{year}{1984}) \bibinfo{pages}{229--259}.
\bibitem[{Koco{\'n} et~al.(2021)Koco{\'n}, Gruza, Bielaniewicz, Grimling, Kanclerz, Mi{\l}kowski, and Kazienko}]{kocon2021learning}
\bibinfo{author}{J.~Koco{\'n}}, \bibinfo{author}{M.~Gruza}, \bibinfo{author}{J.~Bielaniewicz}, \bibinfo{author}{D.~Grimling}, \bibinfo{author}{K.~Kanclerz}, \bibinfo{author}{P.~Mi{\l}kowski}, \bibinfo{author}{P.~Kazienko},
\newblock \bibinfo{title}{Learning personal human biases and representations for subjective tasks in natural language processing},
\newblock in: \bibinfo{booktitle}{2021 IEEE International Conference on Data Mining (ICDM)}, \bibinfo{organization}{IEEE}, \bibinfo{year}{2021}, pp. \bibinfo{pages}{1168--1173}.
\bibitem[{M{\"u}ller et~al.(2006)M{\"u}ller, Carpendale, Bibok, and Racine}]{muller2006subjectivity}
\bibinfo{author}{U.~M{\"u}ller}, \bibinfo{author}{J.~Carpendale}, \bibinfo{author}{M.~Bibok}, \bibinfo{author}{T.~Racine},
\newblock \bibinfo{title}{Subjectivity, identification and differentiation: Key issues in early social development},
\newblock \bibinfo{journal}{Monographs of the Society for Research in Child Development}  (\bibinfo{year}{2006}) \bibinfo{pages}{167--179}.
\bibitem[{Othman et~al.(2015)Othman, Hassan, Moawad, and Idrees}]{othman2015using}
\bibinfo{author}{M.~Othman}, \bibinfo{author}{H.~Hassan}, \bibinfo{author}{R.~Moawad}, \bibinfo{author}{A.~M. Idrees},
\newblock \bibinfo{title}{Using nlp approach for opinion types classifier},
\newblock \bibinfo{journal}{Journal of Computers}  (\bibinfo{year}{2015}).
\bibitem[{Galassi et~al.(2023)Galassi, Ruggeri, Barr{\'o}n-Cede{\~n}o, Alam, Caselli, Kutlu, Stru{\ss}, Antici, Hasanain, K{\"o}hler et~al.}]{galassi2023overview}
\bibinfo{author}{A.~Galassi}, \bibinfo{author}{F.~Ruggeri}, \bibinfo{author}{A.~Barr{\'o}n-Cede{\~n}o}, \bibinfo{author}{F.~Alam}, \bibinfo{author}{T.~Caselli}, \bibinfo{author}{M.~Kutlu}, \bibinfo{author}{J.~M. Stru{\ss}}, \bibinfo{author}{F.~Antici}, \bibinfo{author}{M.~Hasanain}, \bibinfo{author}{J.~K{\"o}hler}, et~al.,
\newblock \bibinfo{title}{Overview of the clef-2023 checkthat! lab: Task 2 on subjectivity in news articles},
\newblock in: \bibinfo{booktitle}{24th Working Notes of the Conference and Labs of the Evaluation Forum, CLEF-WN 2023}, \bibinfo{organization}{CEUR Workshop Proceedings (CEUR-WS. org)}, \bibinfo{year}{2023}, pp. \bibinfo{pages}{236--249}.
\bibitem[{Wiebe et~al.(1999)Wiebe, Bruce, and O'Hara}]{wiebe1999development}
\bibinfo{author}{J.~Wiebe}, \bibinfo{author}{R.~Bruce}, \bibinfo{author}{T.~O'Hara},
\newblock \bibinfo{title}{Development and use of a gold-standard data set for subjectivity classifications},
\newblock in: \bibinfo{booktitle}{Proceedings of the 37th annual meeting of the Association for Computational Linguistics}, \bibinfo{year}{1999}, pp. \bibinfo{pages}{246--253}.
\bibitem[{Nakov et~al.(2016)Nakov, Ritter, Rosenthal, Sebastiani, and Stoyanov}]{nakov2016semeval}
\bibinfo{author}{P.~Nakov}, \bibinfo{author}{A.~Ritter}, \bibinfo{author}{S.~Rosenthal}, \bibinfo{author}{F.~Sebastiani}, \bibinfo{author}{V.~Stoyanov},
\newblock \bibinfo{title}{Semeval-2016 task 4: Sentiment analysis in twitter},
\newblock in: \bibinfo{booktitle}{Proceedings of the 10th international workshop on semantic evaluation (SemEval-2016)}, \bibinfo{year}{2016}, pp. \bibinfo{pages}{1--18}.
\bibitem[{Devlin et~al.(2019)Devlin, Chang, Lee, and Toutanova}]{devlin2019bert}
\bibinfo{author}{J.~Devlin}, \bibinfo{author}{M.-W. Chang}, \bibinfo{author}{K.~Lee}, \bibinfo{author}{K.~Toutanova},
\newblock \bibinfo{title}{Bert: Pre-training of deep bidirectional transformers for language understanding},
\newblock in: \bibinfo{booktitle}{Proceedings of the 2019 Conference of the North American Chapter of the Association for Computational Linguistics: Human Language Technologies, Volume 1 (Long and Short Papers)}, \bibinfo{year}{2019}, pp. \bibinfo{pages}{4171--4186}.
\bibitem[{Balahur et~al.(2009)Balahur, Steinberger, van~der Goot, Pouliquen, and Kabadjov}]{balahur2009opinion}
\bibinfo{author}{A.~Balahur}, \bibinfo{author}{R.~Steinberger}, \bibinfo{author}{E.~van~der Goot}, \bibinfo{author}{B.~Pouliquen}, \bibinfo{author}{M.~Kabadjov},
\newblock \bibinfo{title}{Opinion mining on newspaper quotations},
\newblock in: \bibinfo{booktitle}{Proceedings of the 2009 IEEE/WIC/ACM International Joint Conference on Web Intelligence and Intelligent Agent Technology-Volume 03}, \bibinfo{organization}{IEEE Computer Society}, \bibinfo{year}{2009}, pp. \bibinfo{pages}{523--526}.
\bibitem[{Mihalcea et~al.(2007)Mihalcea, Banea, and Wiebe}]{mihalcea2007learning}
\bibinfo{author}{R.~Mihalcea}, \bibinfo{author}{C.~Banea}, \bibinfo{author}{J.~Wiebe},
\newblock \bibinfo{title}{Learning multilingual subjective language via cross-lingual projections},
\newblock in: \bibinfo{booktitle}{Proceedings of the 45th Annual Meeting of the Association of Computational Linguistics}, \bibinfo{year}{2007}, pp. \bibinfo{pages}{976--983}.
\bibitem[{Barr{\'o}n-Cede{\~{n}}o et~al.(2024)Barr{\'o}n-Cede{\~{n}}o, Alam, Chakraborty, Elsayed, Nakov, Przyby{\l}a, Stru{\ss}, Haouari, Hasanain, Ruggeri, Song, and Suwaileh}]{10.1007/978-3-031-56069-9_62}
\bibinfo{author}{A.~Barr{\'o}n-Cede{\~{n}}o}, \bibinfo{author}{F.~Alam}, \bibinfo{author}{T.~Chakraborty}, \bibinfo{author}{T.~Elsayed}, \bibinfo{author}{P.~Nakov}, \bibinfo{author}{P.~Przyby{\l}a}, \bibinfo{author}{J.~M. Stru{\ss}}, \bibinfo{author}{F.~Haouari}, \bibinfo{author}{M.~Hasanain}, \bibinfo{author}{F.~Ruggeri}, \bibinfo{author}{X.~Song}, \bibinfo{author}{R.~Suwaileh},
\newblock \bibinfo{title}{The clef-2024 checkthat! lab: Check-worthiness, subjectivity, persuasion, roles, authorities, and adversarial robustness},
\newblock in: \bibinfo{editor}{N.~Goharian}, \bibinfo{editor}{N.~Tonellotto}, \bibinfo{editor}{Y.~He}, \bibinfo{editor}{A.~Lipani}, \bibinfo{editor}{G.~McDonald}, \bibinfo{editor}{C.~Macdonald}, \bibinfo{editor}{I.~Ounis} (Eds.), \bibinfo{booktitle}{Advances in Information Retrieval}, \bibinfo{publisher}{Springer Nature Switzerland}, \bibinfo{address}{Cham}, \bibinfo{year}{2024}, pp. \bibinfo{pages}{449--458}.
\bibitem[{Nakov et~al.(2022)Nakov, Barr{'o}n-Cede{~n}o, Da~San~Martino, Alam, M{'\i}guez, Caselli, Kutlu, Zaghouani, Li, Shaar et~al.}]{nakov2022clef}
\bibinfo{author}{P.~Nakov}, \bibinfo{author}{A.~Barr{'o}n-Cede{~n}o}, \bibinfo{author}{G.~Da~San~Martino}, \bibinfo{author}{F.~Alam}, \bibinfo{author}{R.~M{'\i}guez}, \bibinfo{author}{T.~Caselli}, \bibinfo{author}{M.~Kutlu}, \bibinfo{author}{W.~Zaghouani}, \bibinfo{author}{C.~Li}, \bibinfo{author}{S.~Shaar}, et~al.,
\newblock \bibinfo{title}{The clef-2022 checkthat! lab on fighting the covid-19 infodemic and fake news detection},
\newblock in: \bibinfo{booktitle}{European Conference on Information Retrieval}, \bibinfo{organization}{Springer}, \bibinfo{year}{2022}, pp. \bibinfo{pages}{416--428}.
\bibitem[{Alam et~al.(2021)Alam, Dalvi, Shaar, Durrani, Mubarak, Nikolov, Da~San~Martino, Ali, Sajjad, Caselli et~al.}]{alam2021fighting}
\bibinfo{author}{F.~Alam}, \bibinfo{author}{F.~Dalvi}, \bibinfo{author}{S.~Shaar}, \bibinfo{author}{N.~Durrani}, \bibinfo{author}{H.~Mubarak}, \bibinfo{author}{A.~Nikolov}, \bibinfo{author}{G.~Da~San~Martino}, \bibinfo{author}{A.~Ali}, \bibinfo{author}{F.~Sajjad}, \bibinfo{author}{T.~Caselli}, et~al.,
\newblock \bibinfo{title}{Fighting the covid-19 infodemic in social media: a holistic perspective and a call to arms},
\newblock in: \bibinfo{booktitle}{Proceedings of the International AAAI Conference on Web and Social Media}, volume~\bibinfo{volume}{15}, \bibinfo{year}{2021}, pp. \bibinfo{pages}{913--922}.
\bibitem[{Xu et~al.(2019)Xu, Liu, Shu, and Yu}]{xu2019bert}
\bibinfo{author}{H.~Xu}, \bibinfo{author}{B.~Liu}, \bibinfo{author}{L.~Shu}, \bibinfo{author}{P.~S. Yu},
\newblock \bibinfo{title}{Bert post-training for review reading comprehension and aspect-based sentiment analysis},
\newblock in: \bibinfo{booktitle}{Proceedings of the 2019 Conference of the North American Chapter of the Association for Computational Linguistics: Human Language Technologies, Volume 1 (Long and Short Papers)}, \bibinfo{year}{2019}, pp. \bibinfo{pages}{2324--2335}.
\bibitem[{Yu and Jiang(2019)}]{yu2019adapting}
\bibinfo{author}{J.~Yu}, \bibinfo{author}{J.~Jiang},
\newblock \bibinfo{title}{Adapting bert for target-oriented multimodal sentiment classification},
\newblock in: \bibinfo{booktitle}{Proceedings of the 28th International Joint Conference on Artificial Intelligence}, \bibinfo{year}{2019}, pp. \bibinfo{pages}{5408--5414}.

\end{thebibliography}

\end{document}